\pdfoutput=1

\documentclass[11pt]{article}
\usepackage{authblk}
\usepackage{acl}

\usepackage{times}
\usepackage{latexsym}

\usepackage[T1]{fontenc}

\usepackage[utf8]{inputenc}

\usepackage{microtype}
\usepackage{float}
\usepackage{graphicx}
\usepackage{subcaption}
\usepackage{mwe}
\usepackage{xcolor}
\usepackage{booktabs}
\usepackage{paralist}

\usepackage{enumitem}
\usepackage{titlesec}
\usepackage{amsmath}
\usepackage{amsfonts}

\title{{P}roblems with {C}osine as a {M}easure of {E}mbedding {S}imilarity for {H}igh {F}requency {W}ords}

\author[1]{\textbf{Kaitlyn Zhou}}
\author[1]{ \textbf{Kawin Ethayarajh}}
\author[2]{\textbf{Dallas Card}}
\author[1]{\textbf{Dan Jurafsky}}
\affil[1]{Stanford University, \{katezhou, kawin, jurafsky\}@stanford.edu}
\affil[2]{University of Michigan, dalc@umich.edu }

\begin{document}
\maketitle

\begin{abstract}
Cosine similarity of contextual embeddings is used in many NLP tasks (e.g., QA, IR, MT) and metrics (e.g., BERTScore). Here, we uncover systematic ways in which word similarities estimated by cosine over BERT embeddings are understated and trace this effect to training data frequency. We find that relative to human judgements, cosine similarity underestimates the similarity of frequent words with other instances of the same word or other words across contexts, even after controlling for polysemy and other factors. We conjecture that this underestimation of similarity for high frequency words is due to differences in the representational geometry of high and low frequency words and provide a formal argument for the two-dimensional case. 

\end{abstract}

\section{Introduction}
Measuring semantic similarity plays a critical role in numerous NLP tasks like  QA, IR, and MT. Many such metrics are based on the cosine similarity between the contextual embeddings of two words (e.g., BERTScore, MoverScore, BERTR, SemDist; \citealp{kim2021evaluating,zhao2019moverscore,mathur2019putting,zhang2019bertscore}). 
Here, we demonstrate that cosine similarity when used with BERT embeddings is highly sensitive to training data frequency.

The impact of frequency on accuracy and reliability has mostly been studied on \textit{static} word embeddings like word2vec.
Low frequency words have low reliability in neighbor judgements
\citep{hellrich-hahn-2016-bad}, and yield smaller inner products \citep{mimno-thompson-2017-strange} with higher variance \citep{ethayarajh2019towards}. Frequency also correlates with stability (overlap in nearest neighbors) \citep{wendlandt-etal-2018-factors}, and plays a role in word analogies and bias \citep{NIPS2016_a486cd07, Caliskan183, zhao-etal-2018-gender,ethayarajh2019understanding}. 
Similar effects have been found in contextual embeddings, particularly for low-frequency senses, which seem to cause difficulties in WSD performance for BERT and RoBERTa \citep{postma-etal-2016-always, blevins-zettlemoyer-2020-moving, gessler2021bert}. 
Other works have examined how word frequency impacts the similarity of \textit{sentence} embeddings \citep{li2020sentence, jiang2022promptbert}.


While previous work has thus mainly focused on reliability or stability of low frequency words or senses, our work asks: how does frequency impact the semantic similarity of high frequency words?



We find that the cosine of BERT embeddings underestimates the similarity of high frequency words  (to other tokens of the same word or to different words) as compared to human judgements.  In a series of regression studies, we find that this underestimation persists even after controlling for confounders like polysemy, part-of-speech, and lemma.
We conjecture that word frequency induces such distortions via differences in the representational geometry. We introduce new methods for characterizing geometric properties of a word's representation in contextual embedding space, and offer a formal argument for why differences in representational geometry affect cosine similarity measurement in the two-dimensional case.\footnote{Code for this paper can be found at \url{https://github.com/katezhou/cosine\_and\_frequency}}

\section{Effect of Frequency on Cosine Similarity}
To understand the effect of word frequency on cosine between BERT embeddings \citep{devlin-etal-2019-bert}, we first approximate the training data frequency of each word in the BERT pre-training corpus from a combination of the March 1, 2020 Wikimedia Download and counts from BookCorpus \citep{7410368, hartmann-dos-santos-2018-nilc}.\footnote{Additional tools used: \url{https://github.com/IlyaSemenov/wikipedia-word-frequency};
\url{https://github.com/attardi/wikiextractor}} We then consider two datasets that include pairs of words in context with associated human similarity judgements of words: Word-In-Context (WiC) (expert-judged pairs of sentences with a target lemma used in either the same or different WordNet, Wiktionary, or VerbNet senses) and Stanford Contextualized Word Similarity dataset (SCWS) (non-expert judged pairs of sentences annotated with human ratings of the similarity of two target terms). Using datasets with human similarity scores allows us to account for human perceived similarities when measuring the impact of frequency on cosine \citep{pilehvar-camacho-collados-2019-wic, huang-etal-2012-improving}.

\subsection{Study 1: WiC}

\paragraph{Method and Dataset}

The authors of WiC used coarse sense divisions as proxies for words having the same or different meaning and created 5,428\footnote{We used a subset of 5,423 of these examples due to minor spelling differences and availability of frequency data.}
pairs of words in context, labeled as having the same or different meaning:
\begin{compactitem}
    \item same meaning: ``I try to avoid the company of gamblers'' and ``We avoided the ball''
    \item different meaning: ``You must carry your camping gear'' and ``Sound carries well over water''.
\end{compactitem}


To obtain BERT-based similarity measurements, we use \texttt{BERT-base-cased}\footnote{\url{https://huggingface.co/bert-base-cased}} to embed each example, average the representations of the target word over the last four hidden layers, and compute cosine similarity for the pair of representations.\footnote{Out-of-vocabulary words are represented as the average of the subword pieces of the word, following  \citet{pilehvar-camacho-collados-2019-wic} and \citet{blevins-zettlemoyer-2020-moving}; we found that representing OOV words by their first token produced nearly identical results.}

\paragraph{Relation between  frequency and similarity in WiC}
We want to use ordinary least squares regression to measure the effect of word frequency on the cosine similarity of BERT embeddings. First, we split the WiC dataset into examples that were labeled as having the ``same'' or ``different'' meanings. This allows us to to control for perceived similarity of the two words in context --- any frequency effects found within these subsets cannot be explained by variation in human judgements. Next, we control for a number of other confounding factors by including them as variables in our OLS regression. For each target lemma we considered:
\begin{compactitem}
\item[\textbf{frequency}:] $\log_2$ of the number of occurrences in BERT’s training data
\item[\textbf{polysemy}:] $\log_2$ of number of senses in WordNet
\item[\textbf{is\_noun}:] binary indicator for nouns vs. verbs
\item[\textbf{same\_wordform}:] binary indicator of having the same wordform in both contexts (e.g., {\em act}/{\em act} vs.~{\em carry}/{\em carries}) (case insensitive)
\end{compactitem}

\begin{figure}[]
    \includegraphics[width=0.9\linewidth]{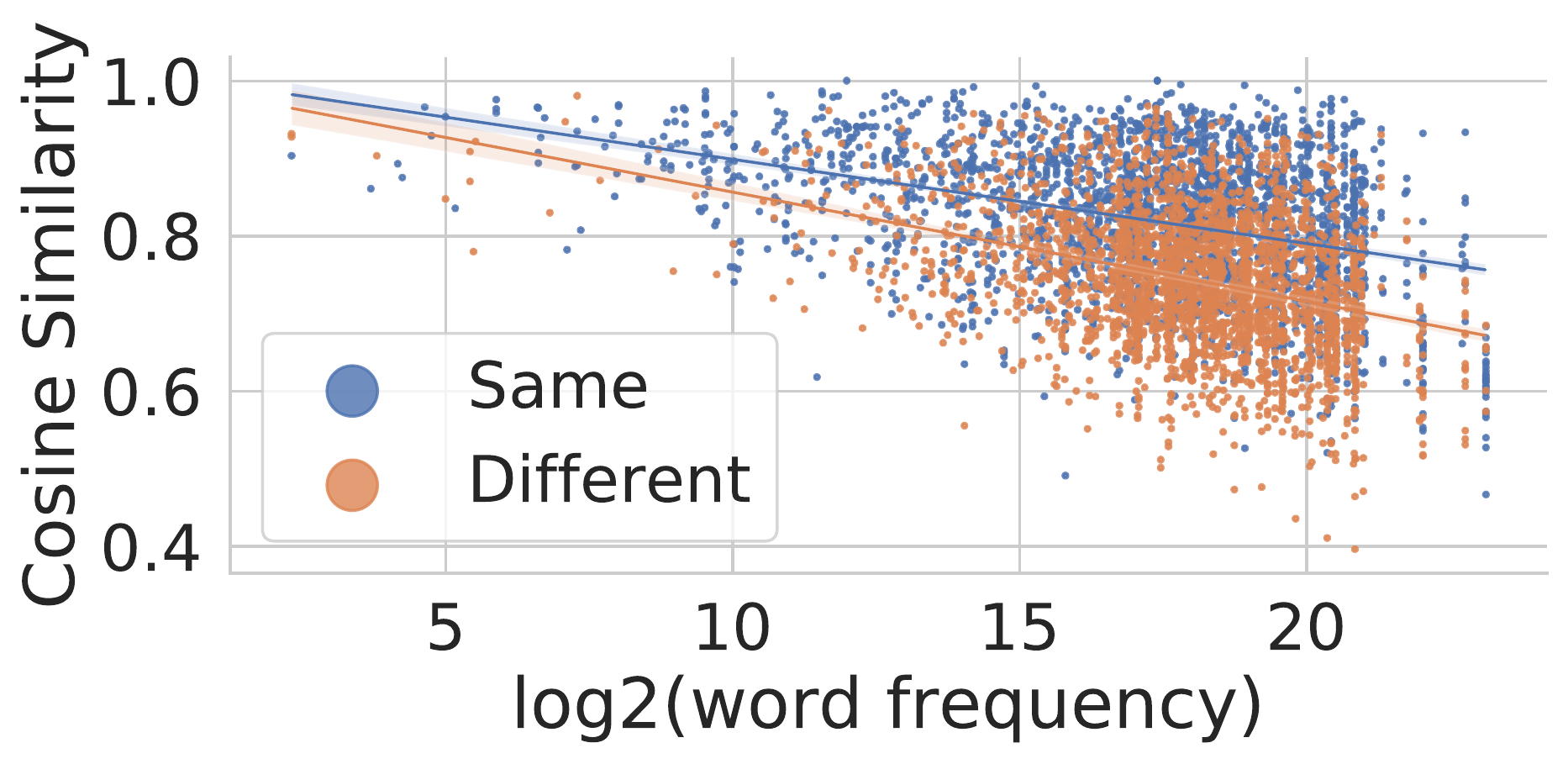}
    \caption{Ordinary Least Squares regression of cosine similarity against frequency, for examples with the same meaning (blue) and different meaning (orange). Both regressions show a significant negative association between cosine similarity and frequency.}
    \label{fig:freq_regression}
\end{figure}

An OLS regression predicting cosine similarity from a single independent factor of $\log_2(\textrm{freq})$ shows a significant negative association between cosine and frequency among "same meaning" examples ($R^2: 0.13$, coeff's $p < 0.001$) and "different meaning" examples ($R^2: 0.14$, coeff's $p < 0.001$) (see Figure \ref{fig:freq_regression}). The same negative frequency effect is found across various model specifications (Table \ref{table:summary1} in Appendix), which also show significantly greater cosine similarity for those examples with the same wordform, a significant negative association with number of senses, and no difference between nouns and verbs. 
In summary, we find that using cosine  to measure the semantic similarity of words via their BERT embeddings gives systematically smaller similarities the higher the frequency of the word.

\paragraph{Results: Comparing to human similarity} To compare cosine similarities to WiC's binary human judgements (same/different meaning), we  followed WiC authors by thresholding cosine values,  tuning the threshold on the training set (resulting threshold: $0.8$). As found in the original WiC paper, cosine similarity is  somewhat predictive of the expert judgements (0.66 dev accuracy, comparable to 0.65 test accuracy from the WiC authors).\footnote{The test set is hidden due to an ongoing leaderboard.} 

\begin{figure}[h!]
  \includegraphics[width=0.9\linewidth, height=5cm]{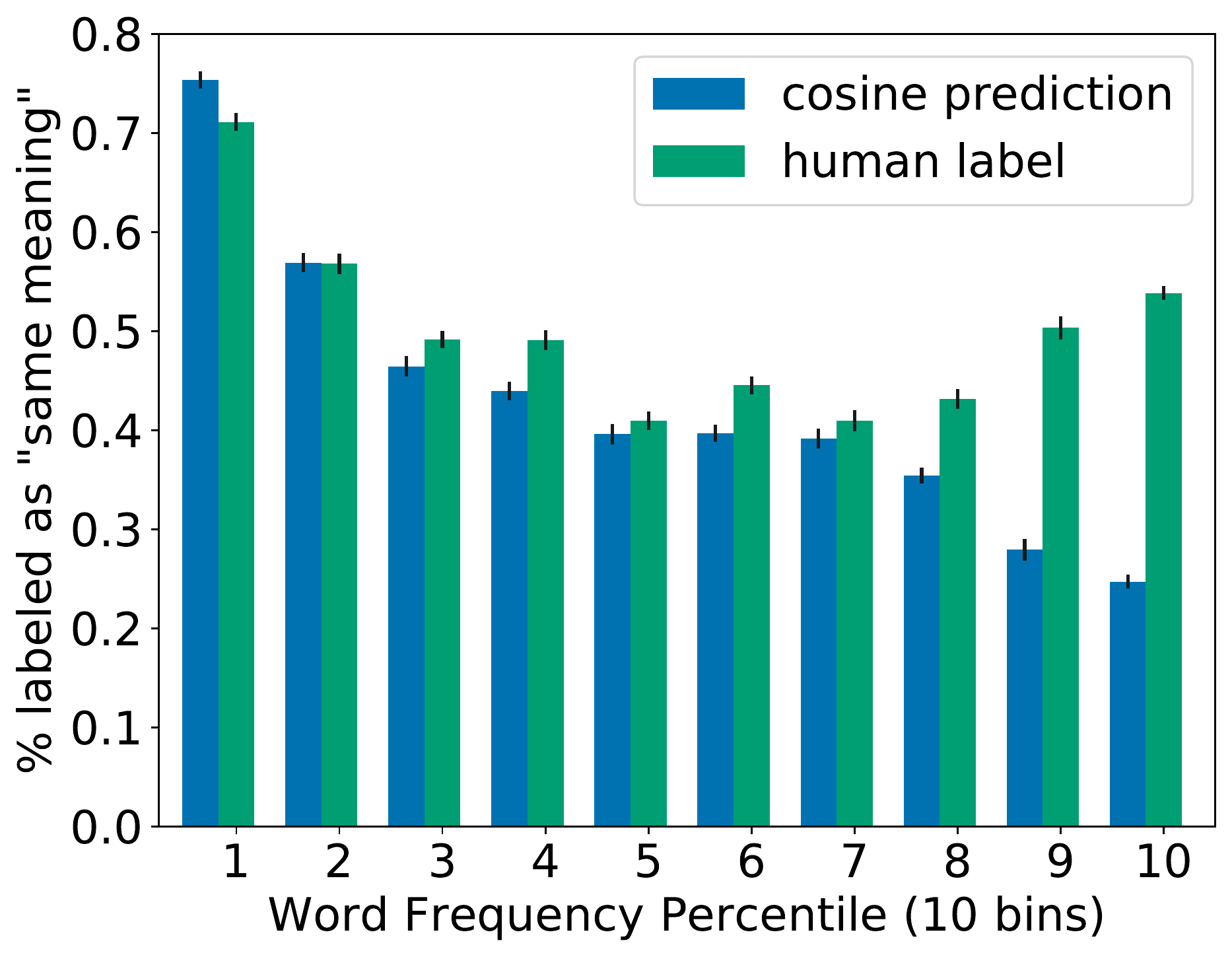}
    \caption{Percentage of examples labeled as having the ``same meaning''. In high frequency words, cosine similarity-based predictions (blue/left) on average \textbf{under}-estimate the similarity of words as compared to human judgements (green/right).}
    \label{fig:cosine_label_vs_frequency}
\end{figure}

Examining the errors as a function of frequency reveals that cosine similarity is a less reliable predictor of human similarity judgements for common terms. Figure \ref{fig:cosine_label_vs_frequency} shows the average proportion of examples predicted to be the same meaning as a function of frequency, grouped into ten bins, each with the same number of examples.
In the highest frequency bin, humans judged 54\% of the examples as having the same meaning compared to only 25\% as judged by cosine similarity. This suggests that in the WiC dataset, relative to humans, the model underestimates the sense similarity for high frequency words.

\subsection{Study 2: SCWS}
Our first study shows that after controlling for sense, cosine will tend to be lower for higher frequency terms. However, the WiC dataset only has binary labels of human judgements, and only indicates similarity between occurrences of the same word. We want to measure if these frequency effects persist across different words and control for more fine-grained human similarity judgements.

\paragraph{Method and Dataset}
SCWS contains crowd judgements of the similarity of two words in context (scale of 1 to 10). We split the dataset based on whether the target words are the same or different ($break/break$ vs $dance/sing$); this both allows us to confirm our results from WiC and also determine whether frequency-based effects exist in  similarity measurements across words.\footnote{For consistency across word embeddings, we only use SCWS examples where the keyword appeared lower-cased in context. We reproduced our results with all SCWS examples and found our findings to be qualitatively the same.}
We use the same embedding method as described for WiC, and again use regression to predict cosine similarities from the following features:
\begin{compactitem}
\item[\textbf{frequency}:] average of $log_2(freq)$ of both words
\item[\textbf{polysemy}:] average of $log_2(sense)$ of both words
\item[\textbf{average rating}:] average rating of semantic similarity as judged by humans on a scale of 1 to 10 (highest).
\end{compactitem}

\paragraph{Results}
If we only use frequency, we find that it mildly explains the variance in cosine similarity both within ($R^2: 0.12$, coeff's $p < 0.001$) and across words  ($R^2: 0.06$, coeff's $p < 0.001$). Adding in human average rating as a feature, frequency is still a significant feature with a negative coefficient. High frequency terms thus tend to have lower cosine similarity scores, even after accounting for human judgements. When using all features, the linear regression models explain 34\% of the total variance in cosine similarity, with frequency still having a significant negative effect (Table \ref{table:summary2} in Appendix). Finally, we verify that for a model with only human ratings, error (true - predicted cosine) is negatively correlated with frequency in held out data (Pearson's $r=-0.18$; $p < 0.01$), indicating an underestimation of cosine in high frequency words (see Figure \ref{fig:cosine_residual} in Appendix).

This finding suggests that using frequency as a feature might help to better match human judgements of similarity. We test this hypothesis by training regression models to predict human ratings, we find that frequency does have a significant positive effect (Table \ref{table:summary3} in Appendix) but the overall improvement over using cosine alone is relatively small ($R^2 = 44.6\%$ vs $ R^2 = 44.3\%$ with or without frequency).
We conclude that the problem of underestimation in cosine similarity cannot be resolved simply by using a linear correction for frequency.

\section{Minimum Bounding Hyperspheres}
\label{section:minimum bounding hyperspheres}
In order to understand why frequency influences cosine similarity, we analyze the geometry of the contextual embeddings.
Unlike static vectors -- where each word type is represented by a single point -- the variation in  contextualized embeddings 
depends on a word's frequency in  training data. We'll call embeddings of a single word type \emph{sibling embeddings} or a \emph{sibling cohort}. To measure variation, we'll use the radius of the smallest hypersphere that contains a set of sibling embeddings (the minimum bounding hypersphere). We tested many ways to measure the space created by high-dimensional vectors. Our results are robust to various other measures of variation, including taking the average, max, or variance of pairwise distance between sibling embeddings, the average norm of sibling embeddings, and taking the PCA of these vectors and calculating the convex hull of sibling embeddings in lower dimensions (see Table \ref{table:other_measures} in the Appendix).
Here we relate frequency to spatial variation, providing both empirical evidence and theoretical intuition.

\begin{figure}[t]
\centering
\includegraphics[width=0.9\columnwidth]{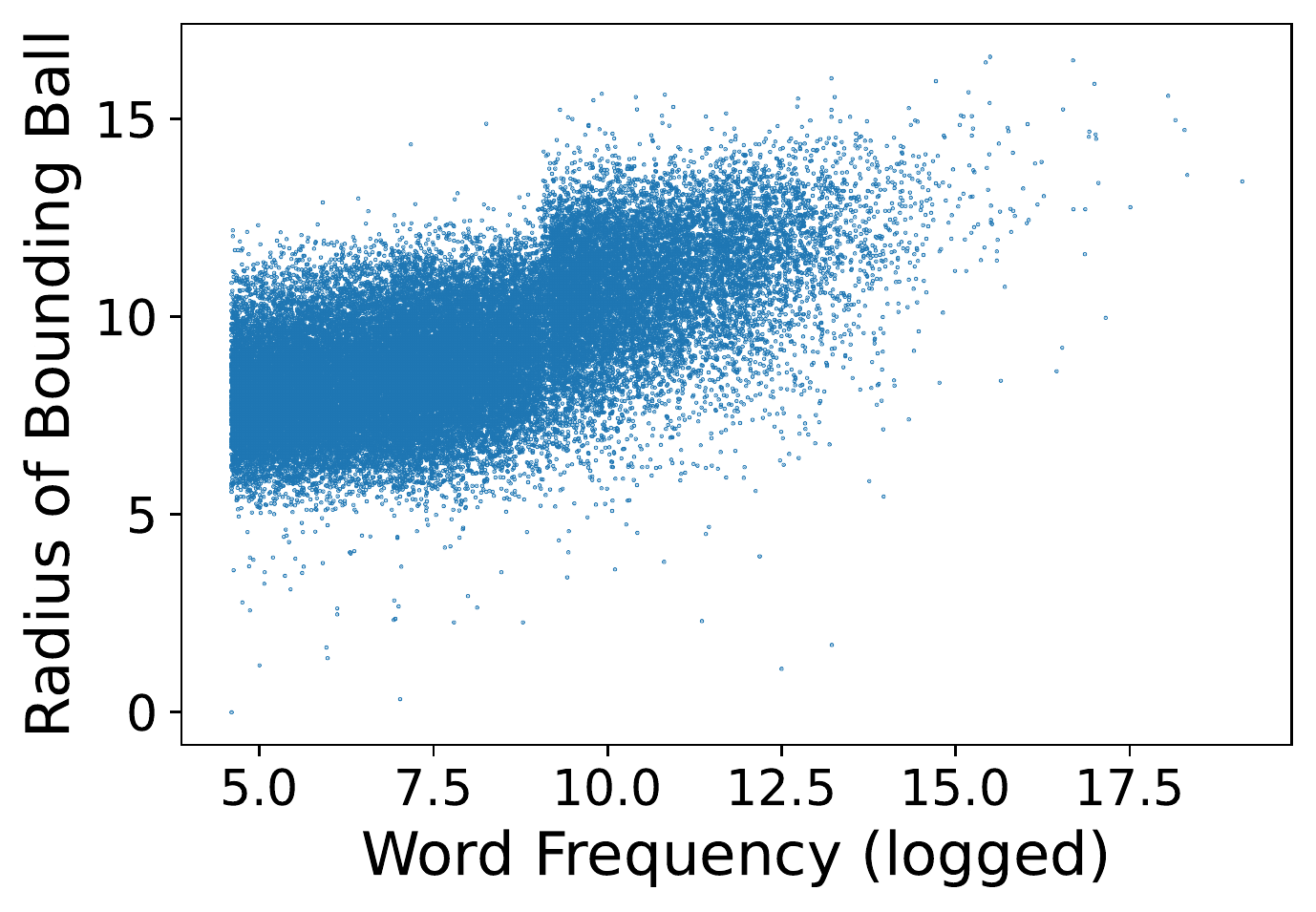}
\caption{The radius of the minimal bounding ball of sibling embeddings of words is correlated with $\log$(word frequency). (Pearson's $r = 0.62$, \textit{p}  < .001)}
\label{fig:radius_frequency}
\end{figure}

For a sample of 39,621 words, for each word we took 10 instances of its sibling embeddings (example sentences queried from Wikipedia), created contexutalized word embeddings using Hugging Face's \texttt{bert-base-cased} model, and calculated the radius of the minimum bounding hypersphere encompassing them.\footnote{Words were binned by frequency and then sampled in order to sample a range of frequencies. As a result, there is a Zipfian effect causing there to be slightly more words in the lower ranges of each bin. We used \url{https://pypi.org/project/miniball/}} \footnote{Given the sensitivity of minimum bounding hypersphere to outliers, we'd imagine that frequency-based distortions would be even more pronounced had we chosen to use more instances of sibling embeddings.}
As shown in Figure \ref{fig:radius_frequency}, there is a significant, strong positive correlation between frequency and size of bounding hypersphere
(Pearson's $r = 0.62, \textit{p} < .001$).
Notably, since the radius was calculated in 768 dimensions, an increase in radius of 1\% results in a hypersphere volume nearly 2084 times larger.\footnote{the n-dimensional volume of a Euclidean ball of radius $R$: $$
V_n(R) = \frac{\pi^{n/2}}{\Gamma(\frac{n}{2} + 1)} R^n$$}

Since frequency and polysemy are highly correlated, we want to measure if frequency is a significant feature for explaining the variance of bounding hyperspheres. Using the unique words of the WiC dataset, we run a series of regressions to predict the radius of bounding hyperspheres. On their own, frequency and polysemy explain for 48\% and 45\% of the radii's variance. Using both features, frequency and polysemy explains for 58\% of the radii's variance and both features are significant -- demonstrating that frequency is a significant feature in predicting radii of bounding hyperspheres (Tables \ref{table:Radius1}, \ref{table:Radius2}, \ref{table:Radius3} in Appendix).

Among the unique words of the WiC dataset, the radii of the target word correlates with training data frequency (Pearson's $r: 0.69, p < 0.001$). Across the WiC dataset, the radii explains for 17\% of the variance in cosine similarity (Table \ref{table:Radius4} in Appendix).\footnote{We used 1,253 out of the original 1,265 unique WiC words and 5,412 out of the original 5,428 WiC examples due to availability of frequency data and contextual examples for target words.}



\subsection{Theoretical Intuition}
Here, we offer some theoretical intuition in 2D for why using cosine similarity to estimate semantic similarity can lead to underestimation (relative to human judgements). Let $\vec{w} \in \mathbb{R}^2$ denote the target word vector, against which we're measuring cosine similarity. Say there were a bounding ball $B_x$ with center $\vec{x_c}$ to which $\vec{w}$ is tangent.
If we normalize every point in the bounding ball, it will form an arc on the unit circle. The length of this arc is $2\theta = 2 \arcsin \frac{r}{\|x_c\|_2}$:
\begin{compactitem}
    \item Let $\theta$ denote the angle made by $x_c$ and the tangent vector $\vec{w}$.
    \item $\sin \theta = \frac{r}{\|x_c\|_2}$, so the arc length on the unit circle is  $r \theta = \arcsin \frac{r}{\|x_c\|_2}$ (normalized points).
    \item  Multiply by 2 to get the arclength between both (normalized) tangent vectors.
\end{compactitem}
Since the arclength is monotonic increasing in $r$, if the bounding ball were larger---while still being tangent to $\vec{w}$---the arclength will be too.

\begin{figure}[t]
\centering
\includegraphics[width=0.7\columnwidth]{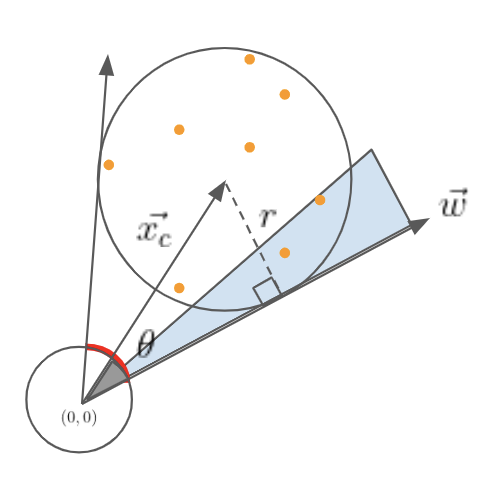}
\caption{An illustration of how using cosine similarity can underestimate word similarity. The cosine similarity between a contextualized representation (orange) and $\vec{w}$ is the dot product of the former's projection onto the red arc of the unit circle (with length $2\theta$) and $\hat{w}$. Only points in the blue region are close enough to $\hat{w}$ to be deemed similar by humans. As the bounding ball grows (e.g., with higher frequency words), if it remains tangent to $\vec{w}$, the fraction of points in the blue region will shrink, leading to underestimation.}
\label{fig:theory}
\end{figure}

The cosine similarity between a point in the bounding ball and $\vec{w}$ is equal to the dot product between the projection of the former onto the unit circle (i.e., somewhere on the arc) and the normalized $\vec{w}$.
This means that only a certain span of the arclength maps to sibling embeddings $\vec{x}_i$ such that $\cos(\vec{x}_i, \vec{w}) \geq t$, where $t$ is the threshold required to be judged as similar by humans (see Footnote 3 and Figure \ref{fig:theory}).
If $B_x$ were larger while still being tangent to $w$, the arclength would increase but the span of the arc containing siblings embeddings sufficiently similar to $w$ would not.
This means a greater proportion of the sibling embeddings will fail to meet this threshold, assuming that the distribution of sibling embeddings in $B_x$ does not change.
Because, in practice, more frequent words have larger bounding balls, depending on how the bounding ball of a word $x$ grows relative to some $\vec{w}$, the similarity of $x$ and $w$ can be underestimated.
This helps explain the findings in Figure \ref{fig:cosine_label_vs_frequency}, but it does not explain why more frequent words have lower similarity with themselves across different contexts, since that requires knowledge of the embedding distribution in the bounding ball. The latter is likely due to more frequent words having less anisotropic representations \citep{ethayarajh2019contextual}.

\section{Discussion and Conclusion}
Cosine distance underestimates compared to humans the semantic similarity of frequent words in a variety of settings (expert versus non-expert judged, and within word sense and across words). This finding has large implications for downstream tasks, given that single-point similarity metrics are used in a variety of methods and experiments \citep{reimers-gurevych-2019-sentence, NEURIPS2019_159c1ffe,zhang2019bertscore, zhao2019moverscore,mathur2019putting,kim2021evaluating}. Word frequency in pre-training data also affects the representational geometry of contextualized embeddings, low frequency words being more concentrated geometrically. One extension of this work might examine how variables such as sentiment and similarity\slash dissimilarity between sentence contexts could impact both human-judged and embedding-based similarity metrics. 

Because training data frequency is something that researchers can control, understanding these  distortions is critical to training large language models. Frequency-based interventions might even be able to correct for these systematic underestimations of similarity (e.g., by modifying training data), which could be important where certain words or subjects may be inaccurately represented. For example, \citet{zhou_richer_countries} illustrates how training data frequencies can lead to discrepancies in the representation of countries, and---since
frequency is highly correlated with a country’s GDP---can perpetuate historic power and wealth inequalities. Future work could also examine how and if frequency effects could be mitigated by post-processing techniques which improve the correlation between human and semantic similarities \citep{timkey-van-schijndel-2021-bark}.

The semantic similarity distortions caused by the over-and under-representation of topics is another reason why documentation for datasets is critical for increasing transparency and accountability in machine learning models \citep{10.1145/3287560.3287596,model_cards, bender-friedman-2018-data, ethayarajh2020utility, ma2021dynaboard}. As language models increase in size and training data becomes more challenging to replicate, we recommend that word frequencies and distortions be revealed to users, bringing awareness to the potential inequalities in datasets and the models that are trained on them. In the future, we hope to see research that more critically examines the downstream implications of these findings and various mitigation techniques for such distortions.

\section*{Acknowledgements}

We sincerely thank Isabel Papadimitriou and our anonymous reviewers for their support, insights, and helpful feedback. This research has been supported in part by a Hoffman-Yee Research Grant from the Stanford Institute for Human-Centered AI, award IIS-2128145 from the NSF, Stanford Data Science, a Stanford Graduate Fellowship, a Facebook Fellowship, and Canada's NSERC.

\bibliography{custom}
\bibliographystyle{acl_natbib}

\clearpage
\appendix

\section{Appendix}
\label{sec:appendix}
For readability, we've summarized the key results from the regressions in \ref{table:summary1} and \ref{table:summary2}. Table \ref{table:summary1} contains results from our WiC experiments where we measure frequency's impact on cosine similarity. We control for human judgements of similarity by splitting the dataset by human labels of "same" and "different" meaning words. The same trends hold for the whole dataset as well.

Table \ref{table:summary2} contains results from the SCWS experiments we measure frequency's impact on cosine similarity within and across word similarities. Similar to the WiC results, we see that frequency does impact cosine similarity, with higher words having lower similarities.

Table \ref{table:summary3} contains results from the SCWS experiments where we measure frequency's impact on human ratings. We see that frequency does not explain human ratings but when used in a model with cosine similarity, frequency has a positive coefficient, indicating it is correcting for the underestimation of cosine similarity.

\begin{table*}[t]
\centering
\small
\begin{tabular}{lcccccccc}
\hline
\multicolumn{9}{c}{OLS predicting cosine similarity} \\ \hline
\multicolumn{1}{c}{WiC} & \multicolumn{4}{c}{Different Sense Meaning} & \multicolumn{4}{c}{Same Sense Meaning} \\ \hline
 & Model 1 & Model 2 & Model 3 & \multicolumn{1}{l}{Model 4} & Model 1 & Model 2 & Model 3 & \multicolumn{1}{l}{Model 4} \\\hline
$log_2(freq)$ & \textbf{-0.014} & \textbf{-0.012} & \textbf{-0.013} & \textbf{-0.013} & \textbf{-0.011} & \textbf{-0.009} & \textbf{-0.009} & \textbf{-0.010} \\
$log_2(sense$) & - & \textbf{-0.012} & \textbf{-0.008} & \textbf{-0.009} & - & \textbf{-0.006} & \textbf{-0.004} & -0.002 \\
same\_wordform & - & - & \textbf{0.045} & \textbf{0.047} & - & - & \textbf{0.059} & \textbf{0.056} \\
is\_noun & - & - & - & -0.006 & - & - & - & \textbf{0.008} \\\hline
$R^2$ & 0.127 & 0.144 & 0.203 & 0.204 & 0.136 & 0.142 & 0.241 & 0.242 \\\hline
Table Number &  \ref{table:A1}&  \ref{table:A2}& \multicolumn{1}{c}{\ref{table:A3}} & \multicolumn{1}{c}{\ref{table:A4}} & \multicolumn{1}{c}{\ref{table:A5}} & \multicolumn{1}{c}{\ref{table:A6}} & \multicolumn{1}{c}{\ref{table:A7}} & \multicolumn{1}{c}{\ref{table:A8}} \\\hline
\end{tabular}
\caption{Coefficients for each of the variables when used in a OLS regression. Bolded numbers are significant. The WiC dataset is split across examples that were rated to have the same or different meaning by experts. Other confounders (polysemy, part-of-speech, word form) were accounted for as features. In model 1, for a word that is twice as frequent, the decrease in cosine similarity will be 0.011.}
\label{table:summary1}
\end{table*}

\begin{table*}[t]
\begin{center}
\small
\begin{tabular}{lcccccccc}
\hline
\multicolumn{1}{c}{SCWS} & \multicolumn{4}{c}{Within Word Examples} & \multicolumn{4}{c}{Across Words Examples} \\\hline
 & Model 1 & Model 2 & Model 3 & Model 4 & Model 1 & Model 2 & Model 3 & Model 4 \\\hline
$log_2(freq)$) & \textbf{-0.020} & - & \textbf{-0.018} & \textbf{-0.016} & \textbf{-0.011} & - & \textbf{-0.008} & \textbf{-0.008} \\
average rating & - & \textbf{0.022} & \textbf{0.021} & \textbf{0.02} & - & \textbf{0.02} & \textbf{0.02} & \textbf{0.02} \\
$log_2(sense)$& - & - & - & \textbf{-0.019} & - & - & - & -0.001 \\\hline
$R^2$& 0.120 & 0.225 & 0.320 & 0.343 & 0.059 & 0.305 & 0.336 & 0.337 \\\hline
Table Number & \ref{table:B1}&  \ref{table:B2}&  \ref{table:B3}& \ref{table:B4} & \ref{table:B5} & \ref{table:B6} & \ref{table:B7} & \ref{table:B8} \\\hline
\end{tabular}
\caption{Coefficients for each of the variables when used in a OLS regression. Bolded numbers are significant. The SCWS dataset is split across examples that use the same (within word) or different (across word) target words. Other con-founders (polysemy and average rating) were accounted for as features. In model 1, for a word that is twice as frequent, the decrease in cosine similarity will be 0.02.}
\label{table:summary2}
\end{center}
\end{table*}

\begin{table*}[t]
\begin{center}
\begin{tabular}{lcccccc} \hline
\multicolumn{6}{c}{OLS Predicting Average Human Rating (Scale of 1 - 10)} \\\hline
\multicolumn{1}{c}{Feature} & Model 1 & Model 2 & Model 3 & Model 4 & Model 5 \\\hline
avg $log_2(freq)$ & -0.057 & - & \textbf{0.099} & - & \textbf{0.076} \\
avg $log_2(sense)$& - & - & -0.0440 & \textbf{-0.134} & \textbf{-0.189} \\
cosine & - & \textbf{16.345} & \textbf{16.665} & \textbf{13.513} & \textbf{13.809} \\
same\_word & - & - & - & \textbf{1.7228} & \textbf{1.687} \\\hline
$R^2$ & 0.002 & 0.404 & 0.408 & 0.443 & 0.446 \\\hline
Table Number & \ref{table:C1} & \multicolumn{1}{c}{\ref{table:C2}} & \ref{table:C3} & \ref{table:C4} &\ref{table:C5} \\\hline    
\end{tabular}
\caption{Coefficients for each of the variables when used in a OLS regression. Bolded numbers are significant. Other con-founders (polysemy, same word) were accounted for as features. In model 5, for a word that is twice as frequent, the increase in human rating will be 0.076. Notice that frequency only becomes a significant as a feature when used with cosine, indicating that it is correcting for an underestimation.}
\label{table:summary3}
\end{center}
\end{table*}

\section{Regression results from WiC experiments}
Tables \ref{table:A1}, \ref{table:A2}, \ref{table:A3}, \ref{table:A4}, \ref{table:A5}, \ref{table:A6}, \ref{table:A7}, \ref{table:A8}.

\begin{table*}[hbt!]
\begin{center}
\begin{tabular}{lclc}
\toprule
\textbf{Dep. Variable:}    & Cosine Similarity & \textbf{  R-squared:         } &     0.127   \\
\textbf{Model:}            &        OLS        & \textbf{  Adj. R-squared:    } &     0.127   \\
\textbf{Method:}           &   Least Squares   & \textbf{  F-statistic:       } &     395.1   \\
\textbf{Date:}             &  Thu, 14 Oct 2021 & \textbf{  Prob (F-statistic):} &  3.55e-82   \\
\textbf{Time:}             &      22:12:38     & \textbf{  Log-Likelihood:    } &    2947.0   \\
\textbf{No. Observations:} &         2713      & \textbf{  AIC:               } &    -5890.   \\
\textbf{Df Residuals:}     &         2711      & \textbf{  BIC:               } &    -5878.   \\
\textbf{Df Model:}         &            1      & \textbf{                     } &             \\
\bottomrule
\end{tabular}
\begin{tabular}{lcccccc}
                    & \textbf{coef} & \textbf{std err} & \textbf{t} & \textbf{P$> |$t$|$} & \textbf{[0.025} & \textbf{0.975]}  \\
\midrule
\textbf{constant}   &       0.9976  &        0.013     &    77.728  &         0.000        &        0.972    &        1.023     \\
\textbf{log2(freq)} &      -0.0141  &        0.001     &   -19.876  &         0.000        &       -0.015    &       -0.013     \\
\bottomrule
\end{tabular}
\begin{tabular}{lclc}
\textbf{Omnibus:}       &  1.261 & \textbf{  Durbin-Watson:     } &    1.952  \\
\textbf{Prob(Omnibus):} &  0.532 & \textbf{  Jarque-Bera (JB):  } &    1.189  \\
\textbf{Skew:}          &  0.044 & \textbf{  Prob(JB):          } &    0.552  \\
\textbf{Kurtosis:}      &  3.053 & \textbf{  Cond. No.          } &     149.  \\
\bottomrule
\end{tabular}
\caption{OLS regression results predicting cosine similarity among "different meaning" senses.}
\label{table:A1}
\end{center}

\end{table*}
\begin{table*}[hbt!]
\begin{center}
\begin{tabular}{lclc}
\toprule
\textbf{Dep. Variable:}    & Cosine Similarity & \textbf{  R-squared:         } &     0.144   \\
\textbf{Model:}            &        OLS        & \textbf{  Adj. R-squared:    } &     0.144   \\
\textbf{Method:}           &   Least Squares   & \textbf{  F-statistic:       } &     228.2   \\
\textbf{Date:}             &  Thu, 14 Oct 2021 & \textbf{  Prob (F-statistic):} &  2.48e-92   \\
\textbf{Time:}             &      22:12:38     & \textbf{  Log-Likelihood:    } &    2973.7   \\
\textbf{No. Observations:} &         2713      & \textbf{  AIC:               } &    -5941.   \\
\textbf{Df Residuals:}     &         2710      & \textbf{  BIC:               } &    -5924.   \\
\textbf{Df Model:}         &            2      & \textbf{                     } &             \\
\bottomrule
\end{tabular}
\begin{tabular}{lcccccc}
                      & \textbf{coef} & \textbf{std err} & \textbf{t} & \textbf{P$> |$t$|$} & \textbf{[0.025} & \textbf{0.975]}  \\
\midrule
\textbf{constant}     &       0.9997  &        0.013     &    78.627  &         0.000        &        0.975    &        1.025     \\
\textbf{log2(freq)}   &      -0.0115  &        0.001     &   -14.624  &         0.000        &       -0.013    &       -0.010     \\
\textbf{log2(senses)} &      -0.0118  &        0.002     &    -7.330  &         0.000        &       -0.015    &       -0.009     \\
\bottomrule
\end{tabular}
\begin{tabular}{lclc}
\textbf{Omnibus:}       &  8.024 & \textbf{  Durbin-Watson:     } &    1.954  \\
\textbf{Prob(Omnibus):} &  0.018 & \textbf{  Jarque-Bera (JB):  } &    9.222  \\
\textbf{Skew:}          &  0.060 & \textbf{  Prob(JB):          } &  0.00994  \\
\textbf{Kurtosis:}      &  3.259 & \textbf{  Cond. No.          } &     153.  \\
\bottomrule
\end{tabular}
\caption{OLS regression results predicting cosine similarity among "different meaning" senses.}
\label{table:A2}
\end{center}

\end{table*}
\begin{table*}[hbt!]
\begin{center}
\begin{tabular}{lclc}
\toprule
\textbf{Dep. Variable:}    & Cosine Similarity & \textbf{  R-squared:         } &     0.203   \\
\textbf{Model:}            &        OLS        & \textbf{  Adj. R-squared:    } &     0.202   \\
\textbf{Method:}           &   Least Squares   & \textbf{  F-statistic:       } &     230.2   \\
\textbf{Date:}             &  Thu, 14 Oct 2021 & \textbf{  Prob (F-statistic):} & 5.14e-133   \\
\textbf{Time:}             &      22:12:38     & \textbf{  Log-Likelihood:    } &    3070.5   \\
\textbf{No. Observations:} &         2713      & \textbf{  AIC:               } &    -6133.   \\
\textbf{Df Residuals:}     &         2709      & \textbf{  BIC:               } &    -6109.   \\
\textbf{Df Model:}         &            3      & \textbf{                     } &             \\
\bottomrule
\end{tabular}
\begin{tabular}{lcccccc}
                        & \textbf{coef} & \textbf{std err} & \textbf{t} & \textbf{P$> |$t$|$} & \textbf{[0.025} & \textbf{0.975]}  \\
\midrule
\textbf{constant}       &       0.9367  &        0.013     &    71.757  &         0.000        &        0.911    &        0.962     \\
\textbf{log2(freq)}     &      -0.0130  &        0.001     &   -16.984  &         0.000        &       -0.015    &       -0.012     \\
\textbf{log2(senses)}   &      -0.0076  &        0.002     &    -4.833  &         0.000        &       -0.011    &       -0.005     \\
\textbf{same\_wordform} &       0.0447  &        0.003     &    14.158  &         0.000        &        0.039    &        0.051     \\
\bottomrule
\end{tabular}
\begin{tabular}{lclc}
\textbf{Omnibus:}       & 13.328 & \textbf{  Durbin-Watson:     } &    1.917  \\
\textbf{Prob(Omnibus):} &  0.001 & \textbf{  Jarque-Bera (JB):  } &   14.587  \\
\textbf{Skew:}          & -0.123 & \textbf{  Prob(JB):          } & 0.000680  \\
\textbf{Kurtosis:}      &  3.261 & \textbf{  Cond. No.          } &     163.  \\
\bottomrule
\end{tabular}
\caption{OLS regression results predicting cosine similarity among "different meaning" senses.}
\label{table:A3}
\end{center}

\end{table*}
\begin{table*}[hbt!]
\begin{center}
\begin{tabular}{lclc}
\toprule
\textbf{Dep. Variable:}    & Cosine Similarity & \textbf{  R-squared:         } &     0.204   \\
\textbf{Model:}            &        OLS        & \textbf{  Adj. R-squared:    } &     0.203   \\
\textbf{Method:}           &   Least Squares   & \textbf{  F-statistic:       } &     173.4   \\
\textbf{Date:}             &  Thu, 14 Oct 2021 & \textbf{  Prob (F-statistic):} & 2.26e-132   \\
\textbf{Time:}             &      22:12:38     & \textbf{  Log-Likelihood:    } &    3071.8   \\
\textbf{No. Observations:} &         2713      & \textbf{  AIC:               } &    -6134.   \\
\textbf{Df Residuals:}     &         2708      & \textbf{  BIC:               } &    -6104.   \\
\textbf{Df Model:}         &            4      & \textbf{                     } &             \\
\bottomrule
\end{tabular}
\begin{tabular}{lcccccc}
                        & \textbf{coef} & \textbf{std err} & \textbf{t} & \textbf{P$> |$t$|$} & \textbf{[0.025} & \textbf{0.975]}  \\
\midrule
\textbf{constant}       &       0.9355  &        0.013     &    71.569  &         0.000        &        0.910    &        0.961     \\
\textbf{log2(freq)}     &      -0.0126  &        0.001     &   -15.858  &         0.000        &       -0.014    &       -0.011     \\
\textbf{log2(senses)}   &      -0.0090  &        0.002     &    -5.030  &         0.000        &       -0.013    &       -0.005     \\
\textbf{same\_wordform} &       0.0467  &        0.003     &    13.760  &         0.000        &        0.040    &        0.053     \\
\textbf{is\_noun}       &      -0.0061  &        0.004     &    -1.629  &         0.103        &       -0.013    &        0.001     \\
\bottomrule
\end{tabular}
\begin{tabular}{lclc}
\textbf{Omnibus:}       & 14.009 & \textbf{  Durbin-Watson:     } &    1.915  \\
\textbf{Prob(Omnibus):} &  0.001 & \textbf{  Jarque-Bera (JB):  } &   15.019  \\
\textbf{Skew:}          & -0.135 & \textbf{  Prob(JB):          } & 0.000548  \\
\textbf{Kurtosis:}      &  3.244 & \textbf{  Cond. No.          } &     164.  \\
\bottomrule
\end{tabular}
\caption{OLS regression results predicting cosine similarity among "different meaning" senses.}
\label{table:A4}
\end{center}

\end{table*}
\begin{table*}[hbt!]
\begin{center}
\begin{tabular}{lclc}
\toprule
\textbf{Dep. Variable:}    & Cosine Similarity & \textbf{  R-squared:         } &     0.136   \\
\textbf{Model:}            &        OLS        & \textbf{  Adj. R-squared:    } &     0.136   \\
\textbf{Method:}           &   Least Squares   & \textbf{  F-statistic:       } &     427.3   \\
\textbf{Date:}             &  Thu, 14 Oct 2021 & \textbf{  Prob (F-statistic):} &  2.94e-88   \\
\textbf{Time:}             &      22:12:38     & \textbf{  Log-Likelihood:    } &    2926.4   \\
\textbf{No. Observations:} &         2710      & \textbf{  AIC:               } &    -5849.   \\
\textbf{Df Residuals:}     &         2708      & \textbf{  BIC:               } &    -5837.   \\
\textbf{Df Model:}         &            1      & \textbf{                     } &             \\
\bottomrule
\end{tabular}
\begin{tabular}{lcccccc}
                    & \textbf{coef} & \textbf{std err} & \textbf{t} & \textbf{P$> |$t$|$} & \textbf{[0.025} & \textbf{0.975]}  \\
\midrule
\textbf{constant}   &       1.0077  &        0.009     &   109.007  &         0.000        &        0.990    &        1.026     \\
\textbf{log2(freq)} &      -0.0109  &        0.001     &   -20.670  &         0.000        &       -0.012    &       -0.010     \\
\bottomrule
\end{tabular}
\begin{tabular}{lclc}
\textbf{Omnibus:}       & 45.476 & \textbf{  Durbin-Watson:     } &    1.977  \\
\textbf{Prob(Omnibus):} &  0.000 & \textbf{  Jarque-Bera (JB):  } &   45.736  \\
\textbf{Skew:}          & -0.298 & \textbf{  Prob(JB):          } & 1.17e-10  \\
\textbf{Kurtosis:}      &  2.778 & \textbf{  Cond. No.          } &     103.  \\
\bottomrule
\end{tabular}
\caption{OLS regression results predicting cosine similarity among "same meaning" senses.}
\label{table:A5}
\end{center}

\end{table*}
\begin{table*}[hbt!]
\begin{center}
\begin{tabular}{lclc}
\toprule
\textbf{Dep. Variable:}    & Cosine Similarity & \textbf{  R-squared:         } &     0.142   \\
\textbf{Model:}            &        OLS        & \textbf{  Adj. R-squared:    } &     0.141   \\
\textbf{Method:}           &   Least Squares   & \textbf{  F-statistic:       } &     224.2   \\
\textbf{Date:}             &  Thu, 14 Oct 2021 & \textbf{  Prob (F-statistic):} &  8.17e-91   \\
\textbf{Time:}             &      22:12:38     & \textbf{  Log-Likelihood:    } &    2935.6   \\
\textbf{No. Observations:} &         2710      & \textbf{  AIC:               } &    -5865.   \\
\textbf{Df Residuals:}     &         2707      & \textbf{  BIC:               } &    -5847.   \\
\textbf{Df Model:}         &            2      & \textbf{                     } &             \\
\bottomrule
\end{tabular}
\begin{tabular}{lcccccc}
                      & \textbf{coef} & \textbf{std err} & \textbf{t} & \textbf{P$> |$t$|$} & \textbf{[0.025} & \textbf{0.975]}  \\
\midrule
\textbf{constant}     &       0.9974  &        0.010     &   104.755  &         0.000        &        0.979    &        1.016     \\
\textbf{log2(freq)}   &      -0.0090  &        0.001     &   -13.270  &         0.000        &       -0.010    &       -0.008     \\
\textbf{log2(senses)} &      -0.0063  &        0.001     &    -4.283  &         0.000        &       -0.009    &       -0.003     \\
\bottomrule
\end{tabular}
\begin{tabular}{lclc}
\textbf{Omnibus:}       & 38.934 & \textbf{  Durbin-Watson:     } &    1.973  \\
\textbf{Prob(Omnibus):} &  0.000 & \textbf{  Jarque-Bera (JB):  } &   39.612  \\
\textbf{Skew:}          & -0.283 & \textbf{  Prob(JB):          } & 2.50e-09  \\
\textbf{Kurtosis:}      &  2.823 & \textbf{  Cond. No.          } &     109.  \\
\bottomrule
\end{tabular}
\caption{OLS regression results predicting cosine similarity among "same meaning" senses.}
\label{table:A6}
\end{center}

\end{table*}
\begin{table*}[hbt!]
\begin{center}
\begin{tabular}{lclc}
\toprule
\textbf{Dep. Variable:}    & Cosine Similarity & \textbf{  R-squared:         } &     0.241   \\
\textbf{Model:}            &        OLS        & \textbf{  Adj. R-squared:    } &     0.240   \\
\textbf{Method:}           &   Least Squares   & \textbf{  F-statistic:       } &     285.7   \\
\textbf{Date:}             &  Thu, 14 Oct 2021 & \textbf{  Prob (F-statistic):} & 4.36e-161   \\
\textbf{Time:}             &      22:12:38     & \textbf{  Log-Likelihood:    } &    3100.7   \\
\textbf{No. Observations:} &         2710      & \textbf{  AIC:               } &    -6193.   \\
\textbf{Df Residuals:}     &         2706      & \textbf{  BIC:               } &    -6170.   \\
\textbf{Df Model:}         &            3      & \textbf{                     } &             \\
\bottomrule
\end{tabular}
\begin{tabular}{lcccccc}
                        & \textbf{coef} & \textbf{std err} & \textbf{t} & \textbf{P$> |$t$|$} & \textbf{[0.025} & \textbf{0.975]}  \\
\midrule
\textbf{constant}       &       0.8928  &        0.011     &    84.562  &         0.000        &        0.872    &        0.914     \\
\textbf{log2(freq)}     &      -0.0092  &        0.001     &   -14.435  &         0.000        &       -0.010    &       -0.008     \\
\textbf{log2(senses)}   &      -0.0035  &        0.001     &    -2.513  &         0.012        &       -0.006    &       -0.001     \\
\textbf{same\_wordform} &       0.0588  &        0.003     &    18.728  &         0.000        &        0.053    &        0.065     \\
\bottomrule
\end{tabular}
\begin{tabular}{lclc}
\textbf{Omnibus:}       & 80.675 & \textbf{  Durbin-Watson:     } &    1.981  \\
\textbf{Prob(Omnibus):} &  0.000 & \textbf{  Jarque-Bera (JB):  } &   87.234  \\
\textbf{Skew:}          & -0.434 & \textbf{  Prob(JB):          } & 1.14e-19  \\
\textbf{Kurtosis:}      &  3.139 & \textbf{  Cond. No.          } &     130.  \\
\bottomrule
\end{tabular}
\caption{OLS regression results predicting cosine similarity among "same meaning" senses.}
\label{table:A7}
\end{center}

\end{table*}
\begin{table*}[hbt!]
\begin{center}
\begin{tabular}{lclc}
\toprule
\textbf{Dep. Variable:}    & Cosine Similarity & \textbf{  R-squared:         } &     0.242   \\
\textbf{Model:}            &        OLS        & \textbf{  Adj. R-squared:    } &     0.241   \\
\textbf{Method:}           &   Least Squares   & \textbf{  F-statistic:       } &     215.8   \\
\textbf{Date:}             &  Thu, 14 Oct 2021 & \textbf{  Prob (F-statistic):} & 6.75e-161   \\
\textbf{Time:}             &      22:12:38     & \textbf{  Log-Likelihood:    } &    3103.2   \\
\textbf{No. Observations:} &         2710      & \textbf{  AIC:               } &    -6196.   \\
\textbf{Df Residuals:}     &         2705      & \textbf{  BIC:               } &    -6167.   \\
\textbf{Df Model:}         &            4      & \textbf{                     } &             \\
\bottomrule
\end{tabular}
\begin{tabular}{lcccccc}
                        & \textbf{coef} & \textbf{std err} & \textbf{t} & \textbf{P$> |$t$|$} & \textbf{[0.025} & \textbf{0.975]}  \\
\midrule
\textbf{constant}       &       0.8952  &        0.011     &    84.424  &         0.000        &        0.874    &        0.916     \\
\textbf{log2(freq)}     &      -0.0096  &        0.001     &   -14.547  &         0.000        &       -0.011    &       -0.008     \\
\textbf{log2(senses)}   &      -0.0022  &        0.002     &    -1.457  &         0.145        &       -0.005    &        0.001     \\
\textbf{same\_wordform} &       0.0560  &        0.003     &    16.512  &         0.000        &        0.049    &        0.063     \\
\textbf{is\_noun}       &       0.0078  &        0.003     &     2.228  &         0.026        &        0.001    &        0.015     \\
\bottomrule
\end{tabular}
\begin{tabular}{lclc}
\textbf{Omnibus:}       & 76.318 & \textbf{  Durbin-Watson:     } &    1.983  \\
\textbf{Prob(Omnibus):} &  0.000 & \textbf{  Jarque-Bera (JB):  } &   82.141  \\
\textbf{Skew:}          & -0.421 & \textbf{  Prob(JB):          } & 1.46e-18  \\
\textbf{Kurtosis:}      &  3.139 & \textbf{  Cond. No.          } &     132.  \\
\bottomrule
\end{tabular}
\caption{OLS regression results predicting cosine similarity among "same meaning" senses.}
\label{table:A8}
\end{center}

\end{table*}
\section{Regression results from SCWS experiments}

Tables \ref{table:B1},  \ref{table:B2},  \ref{table:B3},  \ref{table:B4},  \ref{table:B5},  \ref{table:B6},  \ref{table:B7},  \ref{table:B8}

\begin{table*}[t]
\begin{center}
\begin{tabular}{lclc}
\toprule
\textbf{Dep. Variable:}    & Cosine Similarity & \textbf{  R-squared:         } &     0.120   \\
\textbf{Model:}            &        OLS        & \textbf{  Adj. R-squared:    } &     0.115   \\
\textbf{Method:}           &   Least Squares   & \textbf{  F-statistic:       } &     28.77   \\
\textbf{Date:}             &  Sat, 12 Mar 2022 & \textbf{  Prob (F-statistic):} &  2.12e-07   \\
\textbf{Time:}             &      12:16:53     & \textbf{  Log-Likelihood:    } &    203.87   \\
\textbf{No. Observations:} &          214      & \textbf{  AIC:               } &    -403.7   \\
\textbf{Df Residuals:}     &          212      & \textbf{  BIC:               } &    -397.0   \\
\textbf{Df Model:}         &            1      & \textbf{                     } &             \\
\textbf{Covariance Type:}  &     nonrobust     & \textbf{                     } &             \\
\bottomrule
\end{tabular}
\begin{tabular}{lcccccc}
                   & \textbf{coef} & \textbf{std err} & \textbf{t} & \textbf{P$> |$t$|$} & \textbf{[0.025} & \textbf{0.975]}  \\
\midrule
\textbf{constant}  &       1.0762  &        0.063     &    17.127  &         0.000        &        0.952    &        1.200     \\
\textbf{avg\_freq} &      -0.0196  &        0.004     &    -5.364  &         0.000        &       -0.027    &       -0.012     \\
\bottomrule
\end{tabular}
\begin{tabular}{lclc}
\textbf{Omnibus:}       &  7.823 & \textbf{  Durbin-Watson:     } &    2.040  \\
\textbf{Prob(Omnibus):} &  0.020 & \textbf{  Jarque-Bera (JB):  } &    9.129  \\
\textbf{Skew:}          & -0.307 & \textbf{  Prob(JB):          } &   0.0104  \\
\textbf{Kurtosis:}      &  3.804 & \textbf{  Cond. No.          } &     169.  \\
\bottomrule
\end{tabular}
\caption{OLS regression results predicting cosine similarity among "same" target words}
\label{table:B1}
\end{center}
\end{table*}

\begin{table*}[t]
\begin{center}
\begin{tabular}{lclc}
\toprule
\textbf{Dep. Variable:}    & Cosine Similarity & \textbf{  R-squared:         } &     0.225   \\
\textbf{Model:}            &        OLS        & \textbf{  Adj. R-squared:    } &     0.221   \\
\textbf{Method:}           &   Least Squares   & \textbf{  F-statistic:       } &     61.58   \\
\textbf{Date:}             &  Sat, 12 Mar 2022 & \textbf{  Prob (F-statistic):} &  2.07e-13   \\
\textbf{Time:}             &      12:20:20     & \textbf{  Log-Likelihood:    } &    217.54   \\
\textbf{No. Observations:} &          214      & \textbf{  AIC:               } &    -431.1   \\
\textbf{Df Residuals:}     &          212      & \textbf{  BIC:               } &    -424.3   \\
\textbf{Df Model:}         &            1      & \textbf{                     } &             \\
\textbf{Covariance Type:}  &     nonrobust     & \textbf{                     } &             \\
\bottomrule
\end{tabular}
\begin{tabular}{lcccccc}
                         & \textbf{coef} & \textbf{std err} & \textbf{t} & \textbf{P$> |$t$|$} & \textbf{[0.025} & \textbf{0.975]}  \\
\midrule
\textbf{constant}        &       0.5856  &        0.021     &    28.308  &         0.000        &        0.545    &        0.626     \\
\textbf{average\_rating} &       0.0223  &        0.003     &     7.847  &         0.000        &        0.017    &        0.028     \\
\bottomrule
\end{tabular}
\begin{tabular}{lclc}
\textbf{Omnibus:}       & 31.336 & \textbf{  Durbin-Watson:     } &    2.183  \\
\textbf{Prob(Omnibus):} &  0.000 & \textbf{  Jarque-Bera (JB):  } &   64.374  \\
\textbf{Skew:}          & -0.711 & \textbf{  Prob(JB):          } & 1.05e-14  \\
\textbf{Kurtosis:}      &  5.279 & \textbf{  Cond. No.          } &     25.5  \\
\bottomrule
\end{tabular}
\caption{OLS regression results predicting cosine similarity among "same" target words}
\label{table:B2}
\end{center}

\end{table*}
\begin{table*}[t]
\begin{center}
\begin{tabular}{lclc}
\toprule
\textbf{Dep. Variable:}    & Cosine Similarity & \textbf{  R-squared:         } &     0.320   \\
\textbf{Model:}            &        OLS        & \textbf{  Adj. R-squared:    } &     0.314   \\
\textbf{Method:}           &   Least Squares   & \textbf{  F-statistic:       } &     49.70   \\
\textbf{Date:}             &  Sat, 12 Mar 2022 & \textbf{  Prob (F-statistic):} &  2.06e-18   \\
\textbf{Time:}             &      12:20:20     & \textbf{  Log-Likelihood:    } &    231.56   \\
\textbf{No. Observations:} &          214      & \textbf{  AIC:               } &    -457.1   \\
\textbf{Df Residuals:}     &          211      & \textbf{  BIC:               } &    -447.0   \\
\textbf{Df Model:}         &            2      & \textbf{                     } &             \\
\textbf{Covariance Type:}  &     nonrobust     & \textbf{                     } &             \\
\bottomrule
\end{tabular}
\begin{tabular}{lcccccc}
                         & \textbf{coef} & \textbf{std err} & \textbf{t} & \textbf{P$> |$t$|$} & \textbf{[0.025} & \textbf{0.975]}  \\
\midrule
\textbf{constant}        &       0.8939  &        0.060     &    14.907  &         0.000        &        0.776    &        1.012     \\
\textbf{avg\_freq}       &      -0.0176  &        0.003     &    -5.434  &         0.000        &       -0.024    &       -0.011     \\
\textbf{average\_rating} &       0.0211  &        0.003     &     7.893  &         0.000        &        0.016    &        0.026     \\
\bottomrule
\end{tabular}
\begin{tabular}{lclc}
\textbf{Omnibus:}       & 18.260 & \textbf{  Durbin-Watson:     } &    2.246  \\
\textbf{Prob(Omnibus):} &  0.000 & \textbf{  Jarque-Bera (JB):  } &   27.332  \\
\textbf{Skew:}          & -0.524 & \textbf{  Prob(JB):          } & 1.16e-06  \\
\textbf{Kurtosis:}      &  4.402 & \textbf{  Cond. No.          } &     197.  \\
\bottomrule
\end{tabular}
\caption{OLS regression results predicting cosine similarity among "same" target words}
\label{table:B3}
\end{center}

\end{table*}
\begin{table*}[t]
\begin{center}
\begin{tabular}{lclc}
\toprule
\textbf{Dep. Variable:}    & Cosine Similarity & \textbf{  R-squared:         } &     0.343   \\
\textbf{Model:}            &        OLS        & \textbf{  Adj. R-squared:    } &     0.334   \\
\textbf{Method:}           &   Least Squares   & \textbf{  F-statistic:       } &     36.58   \\
\textbf{Date:}             &  Sat, 12 Mar 2022 & \textbf{  Prob (F-statistic):} &  4.63e-19   \\
\textbf{Time:}             &      12:20:20     & \textbf{  Log-Likelihood:    } &    235.24   \\
\textbf{No. Observations:} &          214      & \textbf{  AIC:               } &    -462.5   \\
\textbf{Df Residuals:}     &          210      & \textbf{  BIC:               } &    -449.0   \\
\textbf{Df Model:}         &            3      & \textbf{                     } &             \\
\textbf{Covariance Type:}  &     nonrobust     & \textbf{                     } &             \\
\bottomrule
\end{tabular}
\begin{tabular}{lcccccc}
                         & \textbf{coef} & \textbf{std err} & \textbf{t} & \textbf{P$> |$t$|$} & \textbf{[0.025} & \textbf{0.975]}  \\
\midrule
\textbf{constant}        &       0.9469  &        0.062     &    15.214  &         0.000        &        0.824    &        1.070     \\
\textbf{avg\_freq}       &      -0.0161  &        0.003     &    -4.983  &         0.000        &       -0.022    &       -0.010     \\
\textbf{average\_rating} &       0.0198  &        0.003     &     7.417  &         0.000        &        0.015    &        0.025     \\
\textbf{avg\_sense}      &      -0.0192  &        0.007     &    -2.711  &         0.007        &       -0.033    &       -0.005     \\
\bottomrule
\end{tabular}
\begin{tabular}{lclc}
\textbf{Omnibus:}       & 13.882 & \textbf{  Durbin-Watson:     } &    2.255  \\
\textbf{Prob(Omnibus):} &  0.001 & \textbf{  Jarque-Bera (JB):  } &   18.177  \\
\textbf{Skew:}          & -0.458 & \textbf{  Prob(JB):          } & 0.000113  \\
\textbf{Kurtosis:}      &  4.095 & \textbf{  Cond. No.          } &     212.  \\
\bottomrule
\end{tabular}
\caption{OLS regression results predicting cosine similarity among "same" target words}
\label{table:B4}
\end{center}

\end{table*}
\begin{table*}[t]
\begin{center}
\begin{tabular}{lclc}
\toprule
\textbf{Dep. Variable:}    & Cosine Similarity & \textbf{  R-squared:         } &     0.059   \\
\textbf{Model:}            &        OLS        & \textbf{  Adj. R-squared:    } &     0.058   \\
\textbf{Method:}           &   Least Squares   & \textbf{  F-statistic:       } &     87.37   \\
\textbf{Date:}             &  Sat, 12 Mar 2022 & \textbf{  Prob (F-statistic):} &  3.41e-20   \\
\textbf{Time:}             &      12:20:20     & \textbf{  Log-Likelihood:    } &    1557.3   \\
\textbf{No. Observations:} &         1406      & \textbf{  AIC:               } &    -3111.   \\
\textbf{Df Residuals:}     &         1404      & \textbf{  BIC:               } &    -3100.   \\
\textbf{Df Model:}         &            1      & \textbf{                     } &             \\
\textbf{Covariance Type:}  &     nonrobust     & \textbf{                     } &             \\
\bottomrule
\end{tabular}
\begin{tabular}{lcccccc}
                   & \textbf{coef} & \textbf{std err} & \textbf{t} & \textbf{P$> |$t$|$} & \textbf{[0.025} & \textbf{0.975]}  \\
\midrule
\textbf{constant}  &       0.7858  &        0.019     &    42.044  &         0.000        &        0.749    &        0.822     \\
\textbf{avg\_freq} &      -0.0106  &        0.001     &    -9.347  &         0.000        &       -0.013    &       -0.008     \\
\bottomrule
\end{tabular}
\begin{tabular}{lclc}
\textbf{Omnibus:}       & 12.804 & \textbf{  Durbin-Watson:     } &    1.683  \\
\textbf{Prob(Omnibus):} &  0.002 & \textbf{  Jarque-Bera (JB):  } &   16.004  \\
\textbf{Skew:}          & -0.130 & \textbf{  Prob(JB):          } & 0.000335  \\
\textbf{Kurtosis:}      &  3.453 & \textbf{  Cond. No.          } &     145.  \\
\bottomrule
\end{tabular}
\caption{OLS regression results predicting cosine similarity among "different" target words}
\label{table:B5}
\end{center}

\end{table*}
\begin{table*}[t]
\begin{center}
\begin{tabular}{lclc}
\toprule
\textbf{Dep. Variable:}    & Cosine Similarity & \textbf{  R-squared:         } &     0.305   \\
\textbf{Model:}            &        OLS        & \textbf{  Adj. R-squared:    } &     0.304   \\
\textbf{Method:}           &   Least Squares   & \textbf{  F-statistic:       } &     614.9   \\
\textbf{Date:}             &  Sat, 12 Mar 2022 & \textbf{  Prob (F-statistic):} & 7.11e-113   \\
\textbf{Time:}             &      12:20:20     & \textbf{  Log-Likelihood:    } &    1770.2   \\
\textbf{No. Observations:} &         1406      & \textbf{  AIC:               } &    -3536.   \\
\textbf{Df Residuals:}     &         1404      & \textbf{  BIC:               } &    -3526.   \\
\textbf{Df Model:}         &            1      & \textbf{                     } &             \\
\textbf{Covariance Type:}  &     nonrobust     & \textbf{                     } &             \\
\bottomrule
\end{tabular}
\begin{tabular}{lcccccc}
                         & \textbf{coef} & \textbf{std err} & \textbf{t} & \textbf{P$> |$t$|$} & \textbf{[0.025} & \textbf{0.975]}  \\
\midrule
\textbf{constant}        &       0.5366  &        0.004     &   150.800  &         0.000        &        0.530    &        0.544     \\
\textbf{average\_rating} &       0.0208  &        0.001     &    24.796  &         0.000        &        0.019    &        0.022     \\
\bottomrule
\end{tabular}
\begin{tabular}{lclc}
\textbf{Omnibus:}       & 32.918 & \textbf{  Durbin-Watson:     } &    1.861  \\
\textbf{Prob(Omnibus):} &  0.000 & \textbf{  Jarque-Bera (JB):  } &   39.508  \\
\textbf{Skew:}          & -0.302 & \textbf{  Prob(JB):          } & 2.64e-09  \\
\textbf{Kurtosis:}      &  3.556 & \textbf{  Cond. No.          } &     8.58  \\
\bottomrule
\end{tabular}
\caption{OLS regression results predicting cosine similarity among "different" target words}
\label{table:B6}
\end{center}

\end{table*}
\begin{table*}[t]
\begin{center}
\begin{tabular}{lclc}
\toprule
\textbf{Dep. Variable:}    & Cosine Similarity & \textbf{  R-squared:         } &     0.336   \\
\textbf{Model:}            &        OLS        & \textbf{  Adj. R-squared:    } &     0.335   \\
\textbf{Method:}           &   Least Squares   & \textbf{  F-statistic:       } &     355.7   \\
\textbf{Date:}             &  Sat, 12 Mar 2022 & \textbf{  Prob (F-statistic):} & 1.12e-125   \\
\textbf{Time:}             &      12:20:20     & \textbf{  Log-Likelihood:    } &    1803.2   \\
\textbf{No. Observations:} &         1406      & \textbf{  AIC:               } &    -3600.   \\
\textbf{Df Residuals:}     &         1403      & \textbf{  BIC:               } &    -3585.   \\
\textbf{Df Model:}         &            2      & \textbf{                     } &             \\
\textbf{Covariance Type:}  &     nonrobust     & \textbf{                     } &             \\
\bottomrule
\end{tabular}
\begin{tabular}{lcccccc}
                         & \textbf{coef} & \textbf{std err} & \textbf{t} & \textbf{P$> |$t$|$} & \textbf{[0.025} & \textbf{0.975]}  \\
\midrule
\textbf{constant}        &       0.6684  &        0.016     &    40.691  &         0.000        &        0.636    &        0.701     \\
\textbf{avg\_freq}       &      -0.0079  &        0.001     &    -8.210  &         0.000        &       -0.010    &       -0.006     \\
\textbf{average\_rating} &       0.0200  &        0.001     &    24.238  &         0.000        &        0.018    &        0.022     \\
\bottomrule
\end{tabular}
\begin{tabular}{lclc}
\textbf{Omnibus:}       & 35.771 & \textbf{  Durbin-Watson:     } &    1.832  \\
\textbf{Prob(Omnibus):} &  0.000 & \textbf{  Jarque-Bera (JB):  } &   44.869  \\
\textbf{Skew:}          & -0.305 & \textbf{  Prob(JB):          } & 1.81e-10  \\
\textbf{Kurtosis:}      &  3.628 & \textbf{  Cond. No.          } &     156.  \\
\bottomrule
\end{tabular}
\caption{OLS regression results predicting cosine similarity among "different" target words}
\label{table:B7}
\end{center}

\end{table*}
\begin{table*}[t]
\begin{center}
\begin{tabular}{lclc}
\toprule
\textbf{Dep. Variable:}    & Cosine Similarity & \textbf{  R-squared:         } &     0.337   \\
\textbf{Model:}            &        OLS        & \textbf{  Adj. R-squared:    } &     0.335   \\
\textbf{Method:}           &   Least Squares   & \textbf{  F-statistic:       } &     237.1   \\
\textbf{Date:}             &  Sat, 12 Mar 2022 & \textbf{  Prob (F-statistic):} & 2.09e-124   \\
\textbf{Time:}             &      12:20:20     & \textbf{  Log-Likelihood:    } &    1803.4   \\
\textbf{No. Observations:} &         1406      & \textbf{  AIC:               } &    -3599.   \\
\textbf{Df Residuals:}     &         1402      & \textbf{  BIC:               } &    -3578.   \\
\textbf{Df Model:}         &            3      & \textbf{                     } &             \\
\textbf{Covariance Type:}  &     nonrobust     & \textbf{                     } &             \\
\bottomrule
\end{tabular}
\begin{tabular}{lcccccc}
                         & \textbf{coef} & \textbf{std err} & \textbf{t} & \textbf{P$> |$t$|$} & \textbf{[0.025} & \textbf{0.975]}  \\
\midrule
\textbf{constant}        &       0.6670  &        0.017     &    40.027  &         0.000        &        0.634    &        0.700     \\
\textbf{avg\_freq}       &      -0.0076  &        0.001     &    -7.044  &         0.000        &       -0.010    &       -0.005     \\
\textbf{average\_rating} &       0.0199  &        0.001     &    23.983  &         0.000        &        0.018    &        0.022     \\
\textbf{avg\_sense}      &      -0.0010  &        0.002     &    -0.516  &         0.606        &       -0.005    &        0.003     \\
\bottomrule
\end{tabular}
\begin{tabular}{lclc}
\textbf{Omnibus:}       & 36.276 & \textbf{  Durbin-Watson:     } &    1.832  \\
\textbf{Prob(Omnibus):} &  0.000 & \textbf{  Jarque-Bera (JB):  } &   45.556  \\
\textbf{Skew:}          & -0.308 & \textbf{  Prob(JB):          } & 1.28e-10  \\
\textbf{Kurtosis:}      &  3.632 & \textbf{  Cond. No.          } &     160.  \\
\bottomrule
\end{tabular}
\caption{OLS regression results predicting cosine similarity among "different" target words}
\label{table:B8}
\end{center}

\end{table*}
\section{Regression results from SCWS experiments, explaining for the difference between cosine similarity and human judgements}
Tables \ref{table:C1}, \ref{table:C2}, \ref{table:C3}, \ref{table:C4}, \ref{table:C5}.

Cosine similarity is partially predictive of human similarity judgements. The full model shows a significant positive effect of frequency \ref{table:C5} indicating that for a given level of cosine similarity, more frequent terms will judged by humans to be more similar, again demonstrating that cosine under-estimates semantic similarity for frequent terms. 

The effect is relatively small, however; for a word that is twice as frequent, the increase in human rating will be 0.0989 (See table \ref{table:C4}). Removing frequency from the model reduces $R^2$ from 40.8\% to 40.4\%. Polysemy shows the opposite effect; those words with more senses are likely to be rated as less similar. In a model with only cosine and polysemy factors, however, frequency has no relationship with human judgements, indicating that including frequency is correcting for the semantic distortion of cosine in the full model.

\begin{table*}[t]
\begin{center}
\begin{tabular}{lclc}
\toprule
\textbf{Dep. Variable:}    &   Human Rating   & \textbf{  R-squared:         } &     0.002   \\
\textbf{Model:}            &       OLS        & \textbf{  Adj. R-squared:    } &     0.001   \\
\textbf{Method:}           &  Least Squares   & \textbf{  F-statistic:       } &     3.074   \\
\textbf{Date:}             & Sat, 12 Mar 2022 & \textbf{  Prob (F-statistic):} &   0.0797    \\
\textbf{Time:}             &     13:15:45     & \textbf{  Log-Likelihood:    } &   -3750.9   \\
\textbf{No. Observations:} &        1620      & \textbf{  AIC:               } &     7506.   \\
\textbf{Df Residuals:}     &        1618      & \textbf{  BIC:               } &     7517.   \\
\textbf{Df Model:}         &           1      & \textbf{                     } &             \\
\textbf{Covariance Type:}  &    nonrobust     & \textbf{                     } &             \\
\bottomrule
\end{tabular}
\begin{tabular}{lcccccc}
                   & \textbf{coef} & \textbf{std err} & \textbf{t} & \textbf{P$> |$t$|$} & \textbf{[0.025} & \textbf{0.975]}  \\
\midrule
\textbf{constant}  &       5.0152  &        0.538     &     9.330  &         0.000        &        3.961    &        6.070     \\
\textbf{avg\_freq} &      -0.0568  &        0.032     &    -1.753  &         0.080        &       -0.120    &        0.007     \\
\bottomrule
\end{tabular}
\begin{tabular}{lclc}
\textbf{Omnibus:}       & 229.333 & \textbf{  Durbin-Watson:     } &    1.972  \\
\textbf{Prob(Omnibus):} &   0.000 & \textbf{  Jarque-Bera (JB):  } &   91.858  \\
\textbf{Skew:}          &   0.385 & \textbf{  Prob(JB):          } & 1.13e-20  \\
\textbf{Kurtosis:}      &   2.124 & \textbf{  Cond. No.          } &     147.  \\
\bottomrule
\end{tabular}
\caption{OLS regression results predicting average human ratings.}
\label{table:C1}
\end{center}

\end{table*}
\begin{table*}[t]
\begin{center}
\begin{tabular}{lclc}
\toprule
\textbf{Dep. Variable:}     &   Human Rating   & \textbf{  R-squared:         } &     0.404   \\
\textbf{Model:}             &       OLS        & \textbf{  Adj. R-squared:    } &     0.403   \\
\textbf{Method:}            &  Least Squares   & \textbf{  F-statistic:       } &     1096.   \\
\textbf{Date:}              & Sat, 12 Mar 2022 & \textbf{  Prob (F-statistic):} & 6.45e-184   \\
\textbf{Time:}              &     13:15:45     & \textbf{  Log-Likelihood:    } &   -3333.6   \\
\textbf{No. Observations:}  &        1620      & \textbf{  AIC:               } &     6671.   \\
\textbf{Df Residuals:}      &        1618      & \textbf{  BIC:               } &     6682.   \\
\textbf{Df Model:}          &           1      & \textbf{                     } &             \\
\textbf{Covariance Type:}   &    nonrobust     & \textbf{                     } &             \\
\bottomrule
\end{tabular}
\begin{tabular}{lcccccc}
                            & \textbf{coef} & \textbf{std err} & \textbf{t} & \textbf{P$> |$t$|$} & \textbf{[0.025} & \textbf{0.975]}  \\
\midrule
\textbf{constant}           &      -6.2058  &        0.314     &   -19.748  &         0.000        &       -6.822    &       -5.589     \\
\textbf{cosine\_similarity} &      16.3453  &        0.494     &    33.101  &         0.000        &       15.377    &       17.314     \\
\bottomrule
\end{tabular}
\begin{tabular}{lclc}
\textbf{Omnibus:}       & 25.721 & \textbf{  Durbin-Watson:     } &    1.974  \\
\textbf{Prob(Omnibus):} &  0.000 & \textbf{  Jarque-Bera (JB):  } &   24.246  \\
\textbf{Skew:}          &  0.260 & \textbf{  Prob(JB):          } & 5.43e-06  \\
\textbf{Kurtosis:}      &  2.703 & \textbf{  Cond. No.          } &     14.7  \\
\bottomrule
\end{tabular}
\caption{OLS regression results predicting average human ratings.}
\label{table:C2}
\end{center}

\end{table*}
\begin{table*}[t]
\begin{center}
\begin{tabular}{lclc}
\toprule
\textbf{Dep. Variable:}     &   Human Rating   & \textbf{  R-squared:         } &     0.408   \\
\textbf{Model:}             &       OLS        & \textbf{  Adj. R-squared:    } &     0.407   \\
\textbf{Method:}            &  Least Squares   & \textbf{  F-statistic:       } &     371.8   \\
\textbf{Date:}              & Sat, 12 Mar 2022 & \textbf{  Prob (F-statistic):} & 1.31e-183   \\
\textbf{Time:}              &     13:15:45     & \textbf{  Log-Likelihood:    } &   -3327.3   \\
\textbf{No. Observations:}  &        1620      & \textbf{  AIC:               } &     6663.   \\
\textbf{Df Residuals:}      &        1616      & \textbf{  BIC:               } &     6684.   \\
\textbf{Df Model:}          &           3      & \textbf{                     } &             \\
\textbf{Covariance Type:}   &    nonrobust     & \textbf{                     } &             \\
\bottomrule
\end{tabular}
\begin{tabular}{lcccccc}
                            & \textbf{coef} & \textbf{std err} & \textbf{t} & \textbf{P$> |$t$|$} & \textbf{[0.025} & \textbf{0.975]}  \\
\midrule
\textbf{constant}           &      -7.9168  &        0.575     &   -13.778  &         0.000        &       -9.044    &       -6.790     \\
\textbf{avg\_freq}          &       0.0989  &        0.028     &     3.473  &         0.001        &        0.043    &        0.155     \\
\textbf{avg\_sense}         &      -0.0440  &        0.048     &    -0.911  &         0.362        &       -0.139    &        0.051     \\
\textbf{cosine\_similarity} &      16.6654  &        0.500     &    33.304  &         0.000        &       15.684    &       17.647     \\
\bottomrule
\end{tabular}
\begin{tabular}{lclc}
\textbf{Omnibus:}       & 25.797 & \textbf{  Durbin-Watson:     } &    1.972  \\
\textbf{Prob(Omnibus):} &  0.000 & \textbf{  Jarque-Bera (JB):  } &   22.821  \\
\textbf{Skew:}          &  0.235 & \textbf{  Prob(JB):          } & 1.11e-05  \\
\textbf{Kurtosis:}      &  2.657 & \textbf{  Cond. No.          } &     252.  \\
\bottomrule
\end{tabular}
\caption{OLS regression results predicting average human ratings.}
\label{table:C3}
\end{center}

\end{table*}
\begin{table*}[t]
\begin{center}
\begin{tabular}{lclc}
\toprule
\textbf{Dep. Variable:}     &   Human Rating   & \textbf{  R-squared:         } &     0.443   \\
\textbf{Model:}             &       OLS        & \textbf{  Adj. R-squared:    } &     0.442   \\
\textbf{Method:}            &  Least Squares   & \textbf{  F-statistic:       } &     428.7   \\
\textbf{Date:}              & Sat, 12 Mar 2022 & \textbf{  Prob (F-statistic):} & 7.28e-205   \\
\textbf{Time:}              &     13:15:45     & \textbf{  Log-Likelihood:    } &   -3278.2   \\
\textbf{No. Observations:}  &        1620      & \textbf{  AIC:               } &     6564.   \\
\textbf{Df Residuals:}      &        1616      & \textbf{  BIC:               } &     6586.   \\
\textbf{Df Model:}          &           3      & \textbf{                     } &             \\
\textbf{Covariance Type:}   &    nonrobust     & \textbf{                     } &             \\
\bottomrule
\end{tabular}
\begin{tabular}{lcccccc}
                            & \textbf{coef} & \textbf{std err} & \textbf{t} & \textbf{P$> |$t$|$} & \textbf{[0.025} & \textbf{0.975]}  \\
\midrule
\textbf{constant}           &      -4.2809  &        0.379     &   -11.310  &         0.000        &       -5.023    &       -3.539     \\
\textbf{avg\_sense}         &      -0.1339  &        0.044     &    -3.012  &         0.003        &       -0.221    &       -0.047     \\
\textbf{cosine\_similarity} &      13.5126  &        0.547     &    24.707  &         0.000        &       12.440    &       14.585     \\
\textbf{same\_word}         &       1.7228  &        0.161     &    10.668  &         0.000        &        1.406    &        2.040     \\
\bottomrule
\end{tabular}
\begin{tabular}{lclc}
\textbf{Omnibus:}       & 24.052 & \textbf{  Durbin-Watson:     } &    2.007  \\
\textbf{Prob(Omnibus):} &  0.000 & \textbf{  Jarque-Bera (JB):  } &   20.099  \\
\textbf{Skew:}          &  0.203 & \textbf{  Prob(JB):          } & 4.32e-05  \\
\textbf{Kurtosis:}      &  2.635 & \textbf{  Cond. No.          } &     46.2  \\
\bottomrule
\end{tabular}
\caption{OLS regression results predicting average human ratings.}
\label{table:C4}
\end{center}

\end{table*}
\begin{table*}[t]
\begin{center}
\begin{tabular}{lclc}
\toprule
\textbf{Dep. Variable:}     &   Human Rating   & \textbf{  R-squared:         } &     0.446   \\
\textbf{Model:}             &       OLS        & \textbf{  Adj. R-squared:    } &     0.444   \\
\textbf{Method:}            &  Least Squares   & \textbf{  F-statistic:       } &     324.7   \\
\textbf{Date:}              & Sat, 12 Mar 2022 & \textbf{  Prob (F-statistic):} & 3.91e-205   \\
\textbf{Time:}              &     13:15:45     & \textbf{  Log-Likelihood:    } &   -3274.5   \\
\textbf{No. Observations:}  &        1620      & \textbf{  AIC:               } &     6559.   \\
\textbf{Df Residuals:}      &        1615      & \textbf{  BIC:               } &     6586.   \\
\textbf{Df Model:}          &           4      & \textbf{                     } &             \\
\textbf{Covariance Type:}   &    nonrobust     & \textbf{                     } &             \\
\bottomrule
\end{tabular}
\begin{tabular}{lcccccc}
                            & \textbf{coef} & \textbf{std err} & \textbf{t} & \textbf{P$> |$t$|$} & \textbf{[0.025} & \textbf{0.975]}  \\
\midrule
\textbf{constant}           &      -5.5590  &        0.600     &    -9.258  &         0.000        &       -6.737    &       -4.381     \\
\textbf{avg\_freq}          &       0.0757  &        0.028     &     2.738  &         0.006        &        0.021    &        0.130     \\
\textbf{avg\_sense}         &      -0.1892  &        0.049     &    -3.881  &         0.000        &       -0.285    &       -0.094     \\
\textbf{cosine\_similarity} &      13.8092  &        0.556     &    24.816  &         0.000        &       12.718    &       14.901     \\
\textbf{same\_word}         &       1.6872  &        0.162     &    10.435  &         0.000        &        1.370    &        2.004     \\
\bottomrule
\end{tabular}
\begin{tabular}{lclc}
\textbf{Omnibus:}       & 24.612 & \textbf{  Durbin-Watson:     } &    2.005  \\
\textbf{Prob(Omnibus):} &  0.000 & \textbf{  Jarque-Bera (JB):  } &   19.555  \\
\textbf{Skew:}          &  0.187 & \textbf{  Prob(JB):          } & 5.67e-05  \\
\textbf{Kurtosis:}      &  2.612 & \textbf{  Cond. No.          } &     285.  \\
\bottomrule
\end{tabular}
\caption{OLS regression results predicting average human ratings.}
\label{table:C5}
\end{center}
\end{table*}
\section{Regression results from minimum bounding hyperspheres}
Using frequency and polysemy to explain for the variability in bounding ball radii.
Tables \ref{table:Radius1}, \ref{table:Radius2}, \ref{table:Radius3}.
Using radius of the bounding ball to explain for the variability of cosine similarity.
Table \ref{table:Radius4}.

\begin{table*}[t]
\begin{center}
\begin{tabular}{lclc}
\toprule
\textbf{Dep. Variable:}    & Radius of Bounding Ball & \textbf{  R-squared:         } &     0.477   \\
\textbf{Model:}            &           OLS           & \textbf{  Adj. R-squared:    } &     0.477   \\
\textbf{Method:}           &      Least Squares      & \textbf{  F-statistic:       } &     1141.   \\
\textbf{Date:}             &     Sat, 12 Mar 2022    & \textbf{  Prob (F-statistic):} & 2.96e-178   \\
\textbf{Time:}             &         15:46:57        & \textbf{  Log-Likelihood:    } &   -2045.0   \\
\textbf{No. Observations:} &            1253         & \textbf{  AIC:               } &     4094.   \\
\textbf{Df Residuals:}     &            1251         & \textbf{  BIC:               } &     4104.   \\
\textbf{Df Model:}         &               1         & \textbf{                     } &             \\
\textbf{Covariance Type:}  &        nonrobust        & \textbf{                     } &             \\
\bottomrule
\end{tabular}
\begin{tabular}{lcccccc}
                    & \textbf{coef} & \textbf{std err} & \textbf{t} & \textbf{P$> |$t$|$} & \textbf{[0.025} & \textbf{0.975]}  \\
\midrule
\textbf{constant}   &       5.5878  &        0.187     &    29.926  &         0.000        &        5.221    &        5.954     \\
\textbf{log2(freq)} &       0.3927  &        0.012     &    33.774  &         0.000        &        0.370    &        0.416     \\
\bottomrule
\end{tabular}
\begin{tabular}{lclc}
\textbf{Omnibus:}       & 15.637 & \textbf{  Durbin-Watson:     } &    2.053  \\
\textbf{Prob(Omnibus):} &  0.000 & \textbf{  Jarque-Bera (JB):  } &   15.928  \\
\textbf{Skew:}          & -0.275 & \textbf{  Prob(JB):          } & 0.000348  \\
\textbf{Kurtosis:}      &  3.052 & \textbf{  Cond. No.          } &     86.0  \\
\bottomrule
\end{tabular}
\caption{OLS regression results predicting radius of bounding ball using frequency}
\label{table:Radius1}
\end{center}

\end{table*}
\begin{table*}[t]
\begin{center}
\begin{tabular}{lclc}
\toprule
\textbf{Dep. Variable:}    & Radius of Bounding Ball & \textbf{  R-squared:         } &     0.448   \\
\textbf{Model:}            &           OLS           & \textbf{  Adj. R-squared:    } &     0.448   \\
\textbf{Method:}           &      Least Squares      & \textbf{  F-statistic:       } &     1015.   \\
\textbf{Date:}             &     Sat, 12 Mar 2022    & \textbf{  Prob (F-statistic):} & 1.25e-163   \\
\textbf{Time:}             &         15:46:57        & \textbf{  Log-Likelihood:    } &   -2078.7   \\
\textbf{No. Observations:} &            1253         & \textbf{  AIC:               } &     4161.   \\
\textbf{Df Residuals:}     &            1251         & \textbf{  BIC:               } &     4172.   \\
\textbf{Df Model:}         &               1         & \textbf{                     } &             \\
\textbf{Covariance Type:}  &        nonrobust        & \textbf{                     } &             \\
\bottomrule
\end{tabular}
\begin{tabular}{lcccccc}
                      & \textbf{coef} & \textbf{std err} & \textbf{t} & \textbf{P$> |$t$|$} & \textbf{[0.025} & \textbf{0.975]}  \\
\midrule
\textbf{constant}     &       9.0630  &        0.093     &    97.878  &         0.000        &        8.881    &        9.245     \\
\textbf{log2(senses)} &       0.9765  &        0.031     &    31.866  &         0.000        &        0.916    &        1.037     \\
\bottomrule
\end{tabular}
\begin{tabular}{lclc}
\textbf{Omnibus:}       & 12.796 & \textbf{  Durbin-Watson:     } &    2.101  \\
\textbf{Prob(Omnibus):} &  0.002 & \textbf{  Jarque-Bera (JB):  } &   13.940  \\
\textbf{Skew:}          & -0.193 & \textbf{  Prob(JB):          } & 0.000940  \\
\textbf{Kurtosis:}      &  3.344 & \textbf{  Cond. No.          } &     8.52  \\
\bottomrule
\end{tabular}
\caption{OLS regression results predicting radius of bounding ball using senses}
\label{table:Radius2}
\end{center}

\end{table*}
\begin{table*}[t]
\begin{center}
\begin{tabular}{lclc}
\toprule
\textbf{Dep. Variable:}    & Radius of Bounding Ball & \textbf{  R-squared:         } &     0.583   \\
\textbf{Model:}            &           OLS           & \textbf{  Adj. R-squared:    } &     0.582   \\
\textbf{Method:}           &      Least Squares      & \textbf{  F-statistic:       } &     872.2   \\
\textbf{Date:}             &     Sat, 12 Mar 2022    & \textbf{  Prob (F-statistic):} & 7.47e-238   \\
\textbf{Time:}             &         15:46:57        & \textbf{  Log-Likelihood:    } &   -1903.7   \\
\textbf{No. Observations:} &            1253         & \textbf{  AIC:               } &     3813.   \\
\textbf{Df Residuals:}     &            1250         & \textbf{  BIC:               } &     3829.   \\
\textbf{Df Model:}         &               2         & \textbf{                     } &             \\
\textbf{Covariance Type:}  &        nonrobust        & \textbf{                     } &             \\
\bottomrule
\end{tabular}
\begin{tabular}{lcccccc}
                      & \textbf{coef} & \textbf{std err} & \textbf{t} & \textbf{P$> |$t$|$} & \textbf{[0.025} & \textbf{0.975]}  \\
\midrule
\textbf{constant}     &       6.0781  &        0.169     &    35.937  &         0.000        &        5.746    &        6.410     \\
\textbf{log2(freq)}   &       0.2581  &        0.013     &    20.071  &         0.000        &        0.233    &        0.283     \\
\textbf{log2(senses)} &       0.5867  &        0.033     &    17.784  &         0.000        &        0.522    &        0.651     \\
\bottomrule
\end{tabular}
\begin{tabular}{lclc}
\textbf{Omnibus:}       & 21.564 & \textbf{  Durbin-Watson:     } &    2.097  \\
\textbf{Prob(Omnibus):} &  0.000 & \textbf{  Jarque-Bera (JB):  } &   23.741  \\
\textbf{Skew:}          & -0.272 & \textbf{  Prob(JB):          } & 6.99e-06  \\
\textbf{Kurtosis:}      &  3.398 & \textbf{  Cond. No.          } &     88.6  \\
\bottomrule
\end{tabular}
\caption{OLS regression results predicting radius of bounding ball using frequency and senses}
\label{table:Radius3}
\end{center}

\end{table*}

\begin{table*}[t]
\begin{center}
\begin{tabular}{lclc}
\toprule
\textbf{Dep. Variable:}          & Cosine Similarity & \textbf{  R-squared:         } &     0.169   \\
\textbf{Model:}                  &        OLS        & \textbf{  Adj. R-squared:    } &     0.169   \\
\textbf{Method:}                 &   Least Squares   & \textbf{  F-statistic:       } &     1103.   \\
\textbf{Date:}                   &  Sat, 12 Mar 2022 & \textbf{  Prob (F-statistic):} & 2.51e-220   \\
\textbf{Time:}                   &      15:54:04     & \textbf{  Log-Likelihood:    } &    5534.8   \\
\textbf{No. Observations:}       &         5412      & \textbf{  AIC:               } & -1.107e+04  \\
\textbf{Df Residuals:}           &         5410      & \textbf{  BIC:               } & -1.105e+04  \\
\textbf{Df Model:}               &            1      & \textbf{                     } &             \\
\textbf{Covariance Type:}        &     nonrobust     & \textbf{                     } &             \\
\bottomrule
\end{tabular}
\begin{tabular}{lcccccc}
                                 & \textbf{coef} & \textbf{std err} & \textbf{t} & \textbf{P$> |$t$|$} & \textbf{[0.025} & \textbf{0.975]}  \\
\midrule
\textbf{Constant}                &       1.1096  &        0.010     &   111.569  &         0.000        &        1.090    &        1.129     \\
\textbf{Radius of Bounding Ball} &      -0.0255  &        0.001     &   -33.215  &         0.000        &       -0.027    &       -0.024     \\
\bottomrule
\end{tabular}
\begin{tabular}{lclc}
\textbf{Omnibus:}       &  1.512 & \textbf{  Durbin-Watson:     } &    1.721  \\
\textbf{Prob(Omnibus):} &  0.470 & \textbf{  Jarque-Bera (JB):  } &    1.543  \\
\textbf{Skew:}          & -0.027 & \textbf{  Prob(JB):          } &    0.462  \\
\textbf{Kurtosis:}      &  2.938 & \textbf{  Cond. No.          } &     109.  \\
\bottomrule
\end{tabular}
\caption{OLS regression results predicting cosine similarity using radius of the bounding ball.}
\label{table:Radius4}
\end{center}
\end{table*}
\section{Other ways of measuring the space of sibling embeddings}
Using a smaller sample of words (10,000 words out of the initial $\sim$39,000 words), we calculate the space occupied by these sibling embeddings using a variety of other metrics. In each metric, we find strong correlations between (log) frequency and the metric in question (see table \ref{table:other_measures}).

\begin{table*}[t]
\begin{center}
\begin{tabular}{lcc}
\\\hline
                                        & Pearson's R & $p$ \\\hline
Average Pairwise Euclidean Distance     & 0.601       & < 0.001   \\
Max Pairwise Euclidean Distance         & 0.584       & < 0.001   \\
Variance of Pairwise Euclidean Distance & 0.292       & < 0.001  \\
Average Norm of Embeddings              & 0.678       & < 0.001  \\
Area of convex hull*                    & 0.603       & < 0.001  \\\hline
\end{tabular}
\end{center}
\caption{Pearson's correlations for numerous other ways of measuring the space occupied by a sibling cohort of ten instances. *To measure the area of a convex hull, we used PCA to projected the embeddings into 2D space and calculated the area. Measuring the convex hull in 768-dimensional space would have required a lot more data (at least 769 samples).}
\label{table:other_measures}
\end{table*}

\section{Residual of Predicted Cosine}
\begin{figure}[!b]
\centering
\includegraphics[width=0.9\columnwidth]{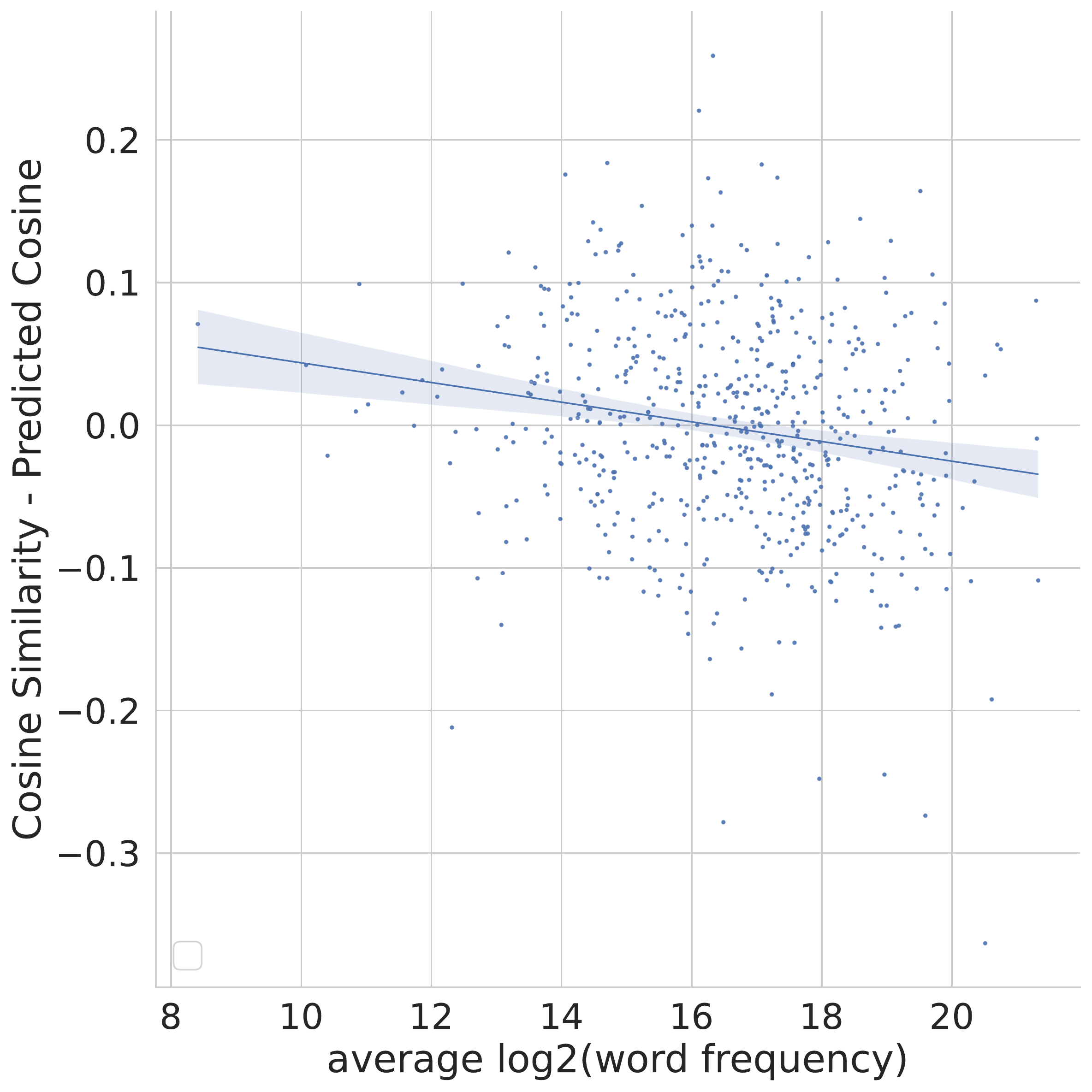}
\caption{Error in cosine similarity and predicted cosine similarity using human ratings. A negative correlation exists, $r = -0.18, p < 0.001$, indicating an underestimation of cosine similarity among the high frequency words.}
\label{fig:cosine_residual}
\end{figure}

For the SCWS dataset, use 1,000 samples as the train set and use the rest as the development set. We train a linear regression model to predict cosine similarity using only human ratings. Taking the difference between cosine similarity and the predicted similarity, we plot this error relative to frequency. We see a negative correlation between this error and frequency $r = -0.18, p < 0.001$, indicating that there is an underestimation of cosine similarity among the high frequency words. Results are shown in Figure \ref{fig:cosine_residual}.

\end{document}